\documentclass[table]{article}

\usepackage{microtype}
\usepackage{graphicx}
\usepackage{subfigure}
\usepackage{booktabs} 

\usepackage{hyperref}

\usepackage[accepted]{icml2024}

\usepackage{amsmath}
\usepackage{amssymb}
\usepackage{mathtools}
\usepackage{amsthm}

\usepackage[capitalize,noabbrev]{cleveref}

\usepackage{xcolor}
\usepackage{array}
\usepackage{multirow}
\usepackage{tabularx}
\usepackage{tcolorbox}
\usepackage{makecell}
\usepackage{bbding}
\usepackage{graphicx}

\theoremstyle{plain}

\theoremstyle{definition}

\theoremstyle{remark}

\usepackage[textsize=tiny]{todonotes}

\icmltitlerunning{Momentor: Advancing Video Large Language Model with Fine-Grained Temporal Reasoning}

\begin{document}

\twocolumn[
\icmltitle{Momentor: Advancing Video Large Language Model\\with Fine-Grained Temporal Reasoning}



\icmlsetsymbol{equal}{*}

\begin{icmlauthorlist}
\icmlauthor{Long Qian}{zju}
\icmlauthor{Juncheng Li}{zju}
\icmlauthor{Yu Wu}{whu}
\icmlauthor{Yaobo Ye}{zju}
\icmlauthor{Hao Fei}{nus}
\icmlauthor{Tat-Seng Chua}{nus}
\icmlauthor{Yueting Zhuang}{zju}
\icmlauthor{Siliang Tang}{zju}
\end{icmlauthorlist}

\icmlaffiliation{zju}{Zhejiang University}
\icmlaffiliation{nus}{National University of Singapore}
\icmlaffiliation{whu}{Wuhan University}

\icmlcorrespondingauthor{Juncheng Li}{junchengli@zju.edu.cn}

\icmlkeywords{Machine Learning, ICML}

\vskip 0.3in
]



\printAffiliationsAndNotice{}  

\vspace{-2mm}
\begin{abstract}
Large Language Models (LLMs) demonstrate remarkable proficiency in comprehending and handling text-based tasks. Many efforts are being made to transfer these attributes to video modality, which are termed Video-LLMs. However, existing Video-LLMs can only capture the coarse-grained semantics and are unable to effectively handle tasks related to comprehension or localization of specific video segments. In light of these challenges, we propose \texttt{Momentor}, a Video-LLM capable of accomplishing fine-grained temporal understanding tasks. To support the training of \texttt{Momentor}, we design an automatic data generation engine to construct \texttt{Moment-10M}, a large-scale video instruction dataset with segment-level instruction data. We train \texttt{Momentor} on \texttt{Moment-10M}, enabling it to perform segment-level reasoning and localization. Zero-shot evaluations on several tasks demonstrate that \texttt{Momentor} excels in fine-grained temporally grounded comprehension and localization. Our project is available at \href{https://github.com/DCDmllm/Momentor}{https://github.com/DCDmllm/Momentor}.
\end{abstract}

\definecolor{myblue}{RGB}{0,173,238}
\definecolor{myred}{RGB}{255,0,0}

\begin{table*}[htbp]
\centering
\resizebox{\textwidth}{!}{
\begin{tabular}{ccccccccccccc}
\hline
Dataset & Total Dur. & Avg Dur. & \#Videos & \#Instructions & \#Segments & \makecell{\#Instances\\Tracks} & \#Actions & \makecell{No Human\\Annotation} & \makecell{Segment-Level\\Comprehension} & \makecell{Temporal\\Localization} & \makecell{Instance\\Reference} &\makecell{Task\\Taxonomy} \\ \hline
VideoChat \cite{videochat} & 41h & 18s & 8.2k & 11.2k & {\color{myred} \XSolidBrush} & {\color{myred} \XSolidBrush} & {\color{myred} \XSolidBrush} & {\color{myblue} \Checkmark} & {\color{myred} \XSolidBrush} & {\color{myred} \XSolidBrush} & {\color{myred} \XSolidBrush} & {\color{myred} \XSolidBrush} \\
Valley \cite{valley} & 608h & 40s & 54.7k & 73.1k & {\color{myred} \XSolidBrush} & {\color{myred} \XSolidBrush} & {\color{myred} \XSolidBrush} & {\color{myblue} \Checkmark} & {\color{myred} \XSolidBrush} & {\color{myred} \XSolidBrush} & {\color{myred} \XSolidBrush} & {\color{myblue} \Checkmark} \\
Video-ChatGPT \cite{video_chatgpt} & 432h & 117s & 13.3k & 100k & {\color{myred} \XSolidBrush} & {\color{myred} \XSolidBrush} & {\color{myred} \XSolidBrush} & {\color{myred} \XSolidBrush} & {\color{myred} \XSolidBrush} & {\color{myred} \XSolidBrush} &{\color{myred} \XSolidBrush} & {\color{myred} \XSolidBrush} \\ \hline
\textbf{\texttt{Moment-10M}} & \textbf{7260h} & \textbf{403s} & \textbf{64.9k} & \textbf{10.4M} & \textbf{1.46M} & \textbf{451.5k} & \textbf{1.51M} & {\color{myblue} \Checkmark} & {\color{myblue} \Checkmark} & {\color{myblue} \Checkmark} & {\color{myblue} \Checkmark} & {\color{myblue} \Checkmark} \\ \hline
\end{tabular}
}
\vspace{-2.5mm}
\caption{Comparison between \texttt{Moment-10M} and existing video instruction datasets}
\vspace{-5mm}
\label{tab:statistics}
\end{table*}

\vspace{-6mm}
\section{Introduction}
\label{submission}
\vspace{-1mm}
Inspired by the success of ChatGPT \cite{chatgpt}, numerous studies across various fields are attempting to integrate Large Language Models (LLMs) with their domain-specific tasks, seeking to bring innovation to these fields. For example, Video Large Language Models (Video-LLMs) such as VideoChat \cite{videochat} and Video-ChatGPT \cite{video_chatgpt} adapt LLM to video modality, striving to merge the understanding, reasoning and interactive skills of LLM with video perception. They typically sample multiple frames from the video, use an image encoder to encode these frames separately, and employ a projection layer (e.g. a linear layer or Q-Former \cite{blip2}) to adapt the visual features to the feature space of an open-source LLM (\cite{llama}, \cite{vicuna}). By training on video-level captioning and QA tasks, they establish coarse-grained multimodal feature alignment and acquire the capability of instruction following.

\begin{figure}[t]
    \centering
    \vspace{-2mm}
    \includegraphics[width=\linewidth]{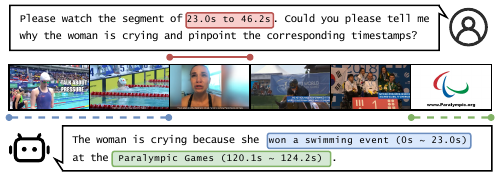}
    \vspace{-8mm}
    \caption{\texttt{Momentor} can perform comprehensive reasoning across multiple segments in a video.}
    \vspace{-6mm}
    \label{fig:illustration}
\end{figure}

Despite being effective, existing Video-LLMs exhibit two limitations: \textbf{(1)~Lack of effective temporal representation.} Existing models encode each sampled frame independently and perform feature projection without retaining precise temporal information in visual features. They lack an effective temporal representation for encoding time positions at inputs and expressing temporal positions accurately at outputs. While directly expressing timestamps in text format seems to be a feasible approach, such a method suffers inherently from precision variability and tokenization complexity of decimals in LLM. \textbf{(2)~Lack of segment-level modeling.} Existing models mainly focus on capturing of global visual semantics, while neglecting the modeling of segment-level semantics and relationships. They are typically trained on trimmed videos (usually around a few seconds) for video-level semantic alignment (video captioning) and instruction-following (video QA). However, common untrimmed videos generally last for several minutes and consist of multiple segments with various contents. Consequently, existing Video-LLMs are unable to provide appropriate responses based on certain segments specified by the user, or locate the segment containing specific content precisely.

To address these challenges, we propose \texttt{Momentor}, a Video-LLM with fine-grained temporal awareness and segment-level reasoning capability. To enhance temporal modeling, we introduce innovations in both model architecture and training methodology. For model architecture, we present \textbf{Temporal Perception Module}, which is designed to flexibly represent accurate temporal positions within videos and inject temporal information into frame features. Temporal Perception Module extends the LLM's vocabulary with a series of temporal tokens designed for temporal positioning and encoding, allowing LLM to precisely perceive fine-grained temporal information and flexibly output accurate timestamps. To avoid the quantization error in representing time with discrete tokens, we incorporate a continuous interpolation mechanism and construct a continuous temporal feature space on top of these temporal tokens. Further, we design a neighboring token propagation mechanism, which propagates the parameter updates of each temporal token to its neighboring tokens to enhance the quality and continuity of the temporal representations. For training, we propose a \textbf{Grounded Event-Sequence Modeling} stage, which trains \texttt{Momentor} to consecutively ground each event in the untrimmed video and caption the corresponding segment with aligned timestamps. Such a temporally grounded event-sequence decoding training bridges the gap between coarse-grained video-level understanding and fine-grained segment-level grounding. It enables \texttt{Momentor} to learn the temporal token space and understand untrimmed videos with complex event sequences.

With fine-grained temporal modeling, we expect that \texttt{Momentor} can learn to perform various segment-level reasoning tasks via instruction tuning. However, existing video instruction datasets do not include segment-level instruction data. Therefore, we propose \texttt{Moment-10M}, a large-scale video instruction fine-tuning dataset with extensive segment-level annotations~(\textsl{e.g.}, actions, tracks). To construct \texttt{Moment-10M}, We design an innovative and automatic data generation engine. Specifically, given a video, we first track all the instances in the video. Then, we design an event boundary detection algorithm to temporally segment the video into coherent events based on video content and instance behaviours. After that, we develop a structured information extraction framework to derive instance, attribute, and event information from the video. We apply a LLM \cite{vicuna} to synthesize these information and generate instruction data. To facilitate comprehensive segment-level reasoning, we design not only \textbf{single-segment tasks} that involve only a single segment, but also \textbf{cross-segment tasks}, which require reasoning over multiple segments to provide correct responses. Employing the data generation engine, we generated 10 million instructions to form \texttt{Moment-10M}. As shown in Table~\ref{tab:statistics}, \texttt{Moment-10M} comprises 1.5 million segments and 451.5 thousand instance tracks while featuring a larger number of videos as well as significantly longer video durations.

We conduct extensive experiments with our proposed \texttt{Momentor}. The results indicate that our \texttt{Momentor} outperforms previous Video-LLMs in multiple tasks involving precise temporal position, such as temporal grounding, dense captioning, action segmentation, and highlight moment retrieval. \texttt{Momentor} demonstrates advanced proficiency in temporal perception. It can provide appropriate responses based on user-indicated segments as well as quickly locate target segments that meet user requirements.

\begin{figure*}[ht]
    \centering
    \includegraphics[width=\textwidth]{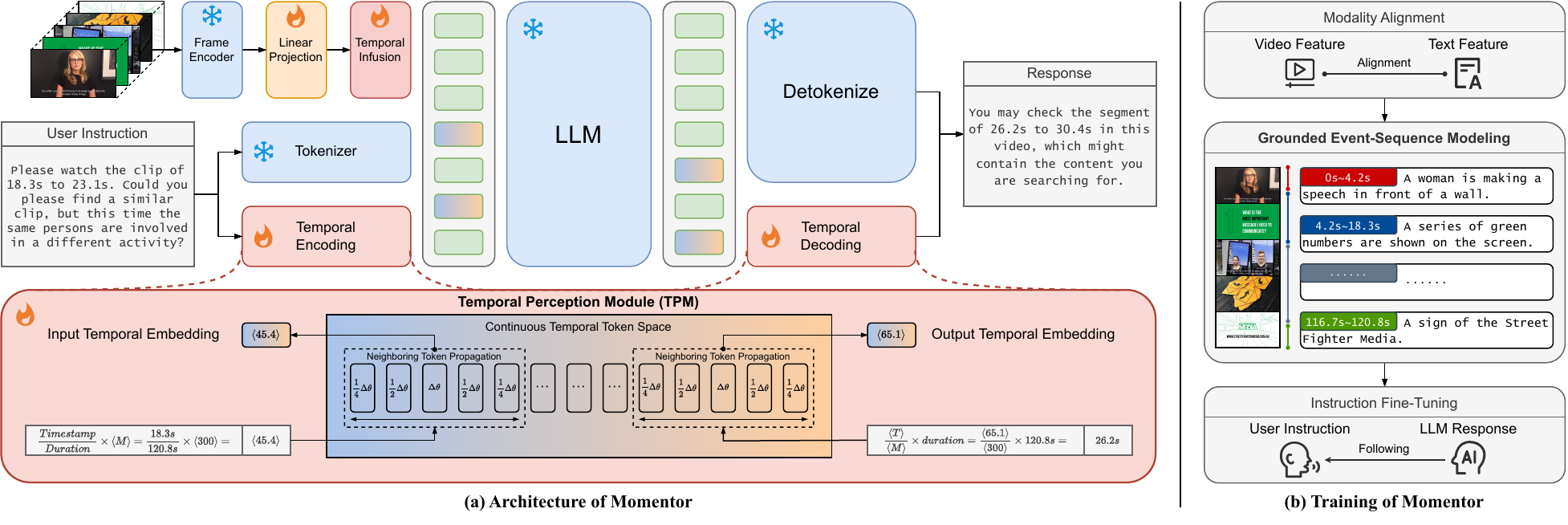}
    \vspace{-9mm}
    \caption{The \textbf{(a)} overall architecture and \textbf{(b)} training of \texttt{Momentor}.}
    \vspace{-3mm}
    \label{fig:momentor}
\end{figure*}

\vspace{-3mm}
\section{Related Work}
\vspace{-1mm}
\subsection{Vision and Language Understanding}
With the rise of deep learning methods in the fields of computer vision and natural language processing, many efforts have been made to explore more complex multimodal understanding of vision and language. For example, tasks such as image and video-based QA, captioning and retrieval have been extensively discussed and explored by many existing studies \cite{vqa, show_and_tell, vse++, ctrl_ret, movieqa, s2vt, dual_enc}. Inspired by the success of the pre-training paradigm in natural language processing and computer vision, many works \cite{clip, blip2, loupe, videobert} propose multimodal pre-trained models with excellent generalization by pre-training on a large amount of image-text or video-text pairs.
\vspace{-1mm}
\subsection{Temporally Grounded Video Understanding}
\vspace{-1mm}
Fine-grained video understanding tasks usually demand the model to view a video as a series of interconnected events and comprehend or locate them in a temporally grounded manner. For instance, \textit{action segmentation} \cite{msbrnn, abd, uvast} requires the model to temporally split the video and output the action label for each segment; \textit{temporal grounding} \cite{charades_sta, 2dtan, vslnet, visa, visa+} demands the model to identify the start and end timestamps of the video segment corresponding to a given natural language query; \textit{highlight moment retrieval} \cite{qvhighlights, univtg} requires the model to find out the central event in a video from a natural language description and pinpoint all related segments; \textit{dense video captioning} \cite{tsp, vid2seq} requires the model to list out all events contained in a video along with their start and end timestamps. Previous methods typically train a task-specific model for each task, whereas we aim to design a unified Video-LLM that can solve these tasks in a zero-shot manner.
\vspace{-3mm}
\subsection{Multimodal Large Language Models}
\vspace{-1mm}
Many efforts have been made to transfer the task-handling capability of Large Language Models (LLMs) to the vision modality, enabling them to complete various tasks based on image content in accordance with user instructions \cite{llava, cheetah, ae_mllm, minigpt4, worldgpt, fact, hyperllava}. Several models \cite{videochat, video_chatgpt, video_llama, valley, vtimellm, timechat} also incorporate temporal information aggregation module with LLM, in order that they can understand video content. Despite being effective in captioning or QA on short videos, the lack of fine-grained temporal modeling in these models prevents them from understanding or locating specific segments in long videos. In contrast, \texttt{Momentor} employs a Temporal Perception Module that integrates a continuous temporal token space for precise temporal positioning and modeling.
\vspace{-4mm}
\section{Momentor}
\vspace{-1mm}
In this section, we present \texttt{Momentor}, a Video-LLM designed for fine-grained comprehension and localization in videos, as shown in Figure~\ref{fig:momentor}. To empower \texttt{Momentor} with fine-grained temporal awareness, we propose the Temporal Perception Module (TPM) (Section \ref{section:tpm}), which facilitates precise temporal positioning and fine-grained temporal information injection. To better train TPM, we introduce Grounded Event-Sequence Modeling (Section \ref{section:gesm}) as an additional pre-training stage, which enables \texttt{Momentor} to comprehend videos in a temporally grounded manner and prepares it for segment-level instruction following tasks.
\vspace{-3mm}
\subsection{Overall Pipeline}
\vspace{-1mm}
\texttt{Momentor} is composed of a frame encoder \cite{ViT}, a linear projection layer, a Temporal Perception Module (TPM), and a Large Language Model (LLM) \cite{llama}. After receiving one input video, \texttt{Momentor} will first uniformly sample multiple frames from the video and encode each frame independently to get frame features. These frame features will be projected into the LLM's feature space by the linear projection layer. The projected features are then processed in the TPM for temporal information injection, which are then concatenated with tokenized user instructions to be the input of LLM. During training, the frame encoder and LLM are kept frozen, while only the linear projection layer and TPM are updated.
\vspace{-3mm}
\subsection{Temporal Perception Module (TPM)}
\label{section:tpm}
\vspace{-1mm}
We propose the Temporal Perception Module to equip \texttt{Momentor} with fine-grained temporal awareness and provide an interface to express precise temporal positions. Specifically, Temporal Perception Module incorporates a continuous temporal token space and employs neighboring token propagation to facilitate the continuity in token space.
\vspace{-9mm}
\paragraph{Continuous Temporal Token Space.} We employ a continuous feature space for precise temporal positioning. Specifically, we uniformly divide the video into $N-1$ segments, and then define $N$ learnable anchor point features to represent the $N-2$ split points and $2$ endpoints, encompassing the relative temporal positions within the video. Then we apply interpolation to define the feature of each temporal point in the timeline, thereby constructing a continuous temporal feature space. With the temporal feature space, we can precisely represent arbitrary temporal positions, enabling \texttt{Momentor} to input or output exact time positions. To unify the training process, we incorporate these anchor point features as specialized temporal tokens into the LLM's vocabulary, denoted as $\langle$1$\rangle$, $\langle$2$\rangle$, ..., $\langle$N$\rangle$, and the outlined feature space is referred as the continuous temporal token space. Therefore, we can train \texttt{Momentor} in an auto-aggressive manner using a unified cross-entropy loss. Studies like \textit{Vid2Seq} \cite{vid2seq} also add specialized tokens to the text decoder's vocabulary to express temporal positions. However, they directly use the discrete tokens for temporal positioning in continuous timelines, which introduces quantization error and prevents them from precise temporal localization. In contrast, our approach solves this problem by constructing a continuous temporal token space on top of these temporal tokens, thereby avoiding quantization error and enabling precise temporal position representation.
\vspace{-3mm}
\paragraph{Neighboring Token Propagation.} 
Unlike language tokens, temporal tokens have a clear sequential relationship. We expect continuity among these temporal tokens, meaning that the embeddings of adjacent tokens should be more similar to each other than those of tokens that are farther apart. However, existing models that use discretized tokens to represent temporal positions have not incorporated any techniques to highlight such continuity. To tackle this issue, we employ a neighboring token propagation mechanism, which enhances continuity by propagating the parameter updates of one temporal token to its adjacent tokens. For any temporal token $\langle$k$\rangle$ involved in the training process, we have:
\vspace{-2mm}
\begin{equation}
\Tilde{t_k}=t_k+t_{adj}-StopGrad(t_{adj}),
\end{equation}
\vspace{-4mm}
\begin{equation}
t_{adj}=\sum_{i=1}^{N}\frac{1}{2^{|i-k|}}\cdot t_i,
\end{equation}
where $\Tilde{t_k}$ is the embedding of temporal token $\langle$k$\rangle$ after neighboring token propagation, $t_i$ is the original embedding for temporal token $\langle$i$\rangle$, $StopGrad$ is the operation to detach a variable's gradient, and $t_{adj}$ is a variable that gathers gradients from all adjacent temporal tokens through a weighted sum. By adding $t_{adj}$ to $t_k$ and subsequently subtracting the gradient-detached $t_{adj}$, we incorporate adjacent temporal tokens into the computation graph, allowing them to receive parameter updates along with $t_k$, while keeping the value of $t_k$ unchanged for precise temporal representation. The weight of each adjacent temporal token in $t_{adj}$ decreases exponentially as their distance to $t_k$ increases. Consequently, temporal tokens closer to $t_k$ receive more similar parameter updates compared to those farther away, and adjacent temporal tokens tend to have more similar embeddings, thereby strengthening the continuity among temporal tokens. We use $\Tilde{t_k}$ instead of $t_k$ in training.
\vspace{-3mm}
\paragraph{Temporal Information Injection.} Since each sampled frame is encoded and projected separately, their features do not contain the corresponding temporal position information. After constructing a continuous temporal token space and applying the neighboring token propagation, now we can actually obtain temporal embeddings corresponding to any timestamp, which contain precise temporal position information and possess the valuable property of temporal continuity. Therefore, we obtain the temporal embeddings at the positions of the sampled frames and directly add them to the projected frame features, as they share the same dimensionality, serving as a form of temporal position encoding to inject fine-grained temporal information.

\begin{figure*}[ht]
    \centering
    \includegraphics[width=\textwidth]{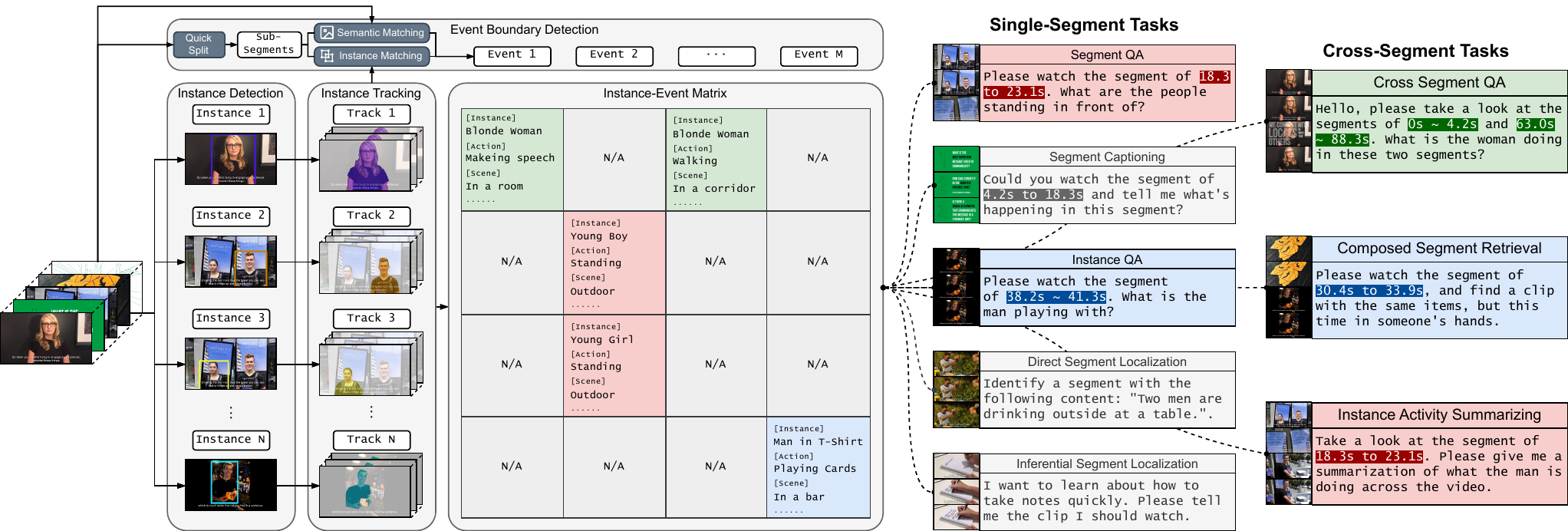}
    \vspace{-7mm}
    \caption{The pipeline of our automatic instruction data generation engine, which can automatically extract structured information from videos and generate diversified instruction data.}
    \vspace{-5mm}
    \label{fig:momentgen}
\end{figure*}

\vspace{-3mm}
\subsection{Grounded Event-Sequence Modeling}
\label{section:gesm}
\vspace{-1mm}
Common untrimmed videos often span several minutes and contain numerous events with diversified content. To facilitate multi-event comprehension, we introduce Grounded Event-Sequence Modeling, an additional pre-training stage focusing on event-sequence decoding, which enables the Temporal Perception Module to align its temporal token space with video timelines and comprehend events in a temporally-grounded manner. We conduct Grounded Event-Sequence Modeling after modality alignment, building temporal awareness upon the aligned multimodal semantics.
\vspace{-6mm}
\paragraph{Modality Alignment.} To align the visual and textual modalities, we train the linear projection layer with a broadly collected dataset of image-text and video-text pairs:
\vspace{-2mm}
\begin{equation}
\mathcal{L}_{align}=-\frac{1}{l}\sum_{i=0}^l\log p(T_C^{i+1}|T_v, T_C^{1:i}),
\end{equation}
where $T_C^i$ is the $i$ th token of the image or video caption $T_C$, and $T_v$ is the frame features.
\vspace{-3mm}
\paragraph{Event-Sequence Decoding.} After the stage of modality alignment, the model only learns the coarse-grained correspondence between visual and textual data. It still lacks fine-grained temporal awareness, so fine-tuning it directly on instruction data with precise timestamps can lead to slow convergence and ineffective event-sequence modeling. Therefore, we apply event-sequence decoding as an intermediary task that bridges the gap between low-level semantic alignment and high-level conceptual interaction. To be precise, given an untrimmed video as input, we require the model to output the event-sequence within it. We represent the $k$ th event as $E_k = [t_{start}^k, t_{end}^k, w_1^k, ..., w_{l_k}^k]$, where $t_{start}^k$, $t_{end}^k$ are the continuous temporal embeddings at the start and end of the $k$ th event, and $[w_1^k, ..., w_{l_k}^k]$ is a general caption composed of $l_{k}$ tokens for this event. The timestamps and general captions of each event in the event-sequence can be conveniently obtained during our instruction generation process without additional calculation (Section~\ref{section:instruction_generation}). We concatenate all the events to formulate the event-sequence $T_E=\{E_i\}_{i=1}^{N_E}$, where $N_E$ is the number of events in the untrimmed video. We apply a language modeling loss for event-sequence decoding:
\vspace{-2mm}
\begin{equation}
\mathcal{L}_{decode} = -\frac{1}{l}\sum_{i=0}^l\log p(T_E^{i+1}|T_v, T_E^{1:i}),
\end{equation}
where $T_E^i$ is the $i$ th token of the event-sequence $T_E$, and $T_v$ is the frame features. With Grounded Event-Sequence Modeling, we establish a preliminary association between the temporal token space and the relative temporal positions within videos, laying the groundwork for segment-level instruction following.
\vspace{-3mm}
\section{Moment-10M}
\vspace{-1mm}
Teaching a Video-LLM to locate specific segments in untrimmed videos and perform complex reasoning on these segments requires substantial training data with fine-grained annotation. However, existing video instruction datasets don't contain instructions with precise timestamps, and their task formats are often limited to captioning, summarizing and basic QA, which overlook the logical associations between events and instances. In light of this, we propose \texttt{Moment-10M}, a large-scale video instruction fine-tuning dataset with segment-level reasoning tasks. To construct \texttt{Moment-10M}, we design a data generation engine that can automatically extract instance and event information along with their relationships from the videos, and then generate corresponding instruction data based on these information, as shown in Figure~\ref{fig:momentgen}. We meticulously design various types of instruction-following tasks, aiming to enhance \texttt{Momentor} in comprehensive segment-level reasoning.
\vspace{-3mm}
\subsection{Structured Information Extraction}
\label{section:information_extraction}
\vspace{-1mm}
The relationships between instances and events in an untrimmed video can be extremely complex. A particular instance might appear in different events that are far apart, and an event might contain several instances that seem unrelated. To fully explore the associations between instances and events within a video, we propose an Event Boundary Detection algorithm that can accurately detect the event boundaries in the video based on the instance information and video content. We then construct an Instance-Event Matrix, to extract and organize the visual information in a structured way, where the spatio-temporal correspondences from a video can be effectively captured.

\begin{table*}[htbp]
\centering
\resizebox{\textwidth}{!}{
\begin{tabular}{|c|cccc|cccc|ccc|}
\hline
\multirow{3}{*}{Model} & \multicolumn{8}{c|}{Action Segmentation} & \multicolumn{3}{c|}{Dense Video Captioning} \\
\cline{2-12}
    & \multicolumn{4}{c|}{Breakfast} & \multicolumn{4}{c|}{50Salads} & \multicolumn{3}{c|}{ActivityNet-Captions} \\
    & MoF & \multicolumn{3}{c|}{F1@\{10, 25, 50\}} & MoF & \multicolumn{3}{c|}{F1@\{10, 25, 50\}} & SODA\_c & CIDEr & METEOR \\
\hline
Video-ChatGPT (7B) \cite{video_chatgpt} & 5.1 & 7.8 & 2.4 & 0.5 & 9.6 & 7.1 & 3.1 & 1.1 & 0.4 & 2.1 & 0.7 \\
VideoChat (7B) \cite{videochat} & 7.9 & 8.8 & 5.3 & 2.8 & 13.3 & 10.6 & 3.5 & 1.1 & 0.7 & 3.3 & 1.2 \\
Video-LLaMA (7B) \cite{video_llama} & 11.6 & 15.2 & 8.8 & 4.2 & 14.3 & 12.9 & 4.0 & 1.2 & 0.9 & 4.6 & 2.4 \\
Valley (7B) \cite{valley} & 4.1 & 7.4 & 4.5 & 2.4 & 13.2 & 11.3 & 3.5 & 1.8 & 0.3 & 1.8 & 0.8 \\
\hline
\textbf{\texttt{Momentor} (7B)} & \textbf{24.4} & \textbf{41.2} & \textbf{33.6} & \textbf{21.8} & \textbf{17.8} & \textbf{22.8} & \textbf{15.9} & \textbf{13.0}  & \textbf{2.3} & \textbf{14.9} & \textbf{4.7} \\
\hline
\end{tabular}
}
\vspace{-3mm}
\caption{Comparison with existing Video-LLMs on dense video captioning and action segmentation}
\vspace{-3mm}
\label{tab:exp_res1}
\end{table*}

\begin{table*}[htbp]
\centering
\resizebox{\textwidth}{!}{
\begin{tabular}{|c|cccc|cccc|cc|}
\hline
\multirow{3}{*}{Model} & \multicolumn{8}{c|}{Temporal Grounding} & \multicolumn{2}{c|}{Highlight Moment Retrieval}\\
\cline{2-11}
& \multicolumn{4}{c|}{ActivityNet-Captions} & \multicolumn{4}{c|}{Charades-STA} & \multicolumn{2}{c|}{QVHighlights} \\
& R@0.3 & R@0.5 & R@0.7 & mIoU & R@0.3 & R@0.5 & R@0.7 & mIoU & mAP & R1@0.5 \\
\hline
Video-ChatGPT (7B) \cite{video_chatgpt} & 19.5 & 10.6 & 4.8 & 14.2 & 27.2 & 6.2 & 1.9 & 19.7 & 3.8 & 8.7 \\
VideoChat (7B) \cite{videochat} & 23.5 & 12.6 & 6.0 & 17.4 & 32.8 & 8.6 & 0.0 & 25.9 & 4.1 & 7.0 \\
Video-LLaMA (7B) \cite{video_llama} & 21.9 & 10.8 & 4.9 & 16.5 & 25.2 & 10.6 & 3.4 & 16.8 & 2.1 & 6.6 \\
Valley (7B) \cite{valley} & 30.6 & 13.7 & 8.1 & 21.9 & 28.4 & 1.8 & 0.3 & 21.4 & 5.3 & 8.7 \\
\hline
\textbf{\texttt{Momentor} (7B)} & \textbf{42.9} & \textbf{23.0} & \textbf{12.4} & \textbf{29.3} & \textbf{42.6} & \textbf{26.6} & \textbf{11.6} & \textbf{28.5} & \textbf{7.6} & \textbf{17.0} \\
\hline
\end{tabular}
}
\vspace{-3mm}
\caption{Comparison with existing Video-LLMs on temporal grounding and highlight moment retrieval}
\vspace{-5mm}
\label{tab:exp_res2}
\end{table*}

\vspace{-4mm}
\paragraph{Event Boundary Detection.} For an arbitrary video to be processed, we first uniformly sample multiple frames from the video. We employ Grounding DINO \cite{grounding_dino} to extract instance information from these sampled frames, and then compare and merge the instances across the sampled frames to obtain the spatio-temporal trajectories of instances in the video, termed as instance tracks. The instance tracks show the dynamics of each instance over time, which also reflect the event transitions in the video. Based on video content and instance dynamics, we design a comprehensive event boundary detection method. We first use PySceneDetect \cite{pyscenedetect} to calculate frame-by-frame differences in the video, resulting in an array of frame difference scores. Then, we apply a Gaussian filter to reduce noise and smooth these scores. We select local maxima that are higher than a certain threshold as split points, to divide the video into several sub-segments. Since such segmentation only considers changes in RGB values and doesn't account for semantic transitions, we adopt a semantic-based merging algorithm to merge adjacent sub-segments that experience abrupt visual changes but still belong to the same event. To be precise, for the two adjacent sub-segments, we extract the last frame from the previous sub-segment and the first frame from the next sub-segment and calculate their consistency value as:
\vspace{-2mm}
\begin{equation}
\begin{split}
& Consistency = \cos(F^{'}, F^{''}) \\
& + \frac{1}{|U_I|} \sum_{i=1}^{|U_I|}\cos(F_{I_i}^{'}, F_{I_i}^{''})\cdot (1-Dist(I_i^{'}, I_i^{''})),
\end{split}
\end{equation}
where $F^{'}$ and $F^{''}$ are visual features of the last frame in the previous sub-segment and the first frame in the next sub-segment, and $U_I$ is the union of instances shown in these two frames. $F_{I_i}^{'}$ and $F_{I_i}^{''}$ are ROI aligned \cite{roi_align} features of the $i$ th instance, and $Dist(I_i^{'}, I_i^{''})$ is the normalized distance between the positions of the $i$ th instance in these two frames with a value in $[0, 1]$. We set this distance to be 1 if the $i$ th instance appears in only one of these two frames. All visual features involved have been obtained during object detection, thus not incurring additional computational costs. We merge two adjacent sub-segments if their consistency value is higher than a set threshold. Consequently, we obtain a series of segments with semantic consistency, each encompassing a coherent event.
\vspace{-3mm}
\paragraph{Instance-Event Matrix.} Based on the result of instance tracking and event segmentation, we construct an Instance-Event Matrix, where each row represents an instance track (the video itself also counts as a track), and each column represents an event. The instance-event matrix shares certain similarities with video scene graphs \cite{video_rel_det, pano_video_scene_graph} as both involve instance behaviour tracking and structured semantic representation, but the instance-event matrix places greater emphasis on modeling the complex associations between events. We traverse the matrix and utilize several multimodal pre-trained models to extract visual clues such as scenes, instances, actions and attributes from each track. With the structured information organized in instance-event matrix, we can quickly generate instruction data that includes various spatio-temporal associations.

\begin{figure}[t]
    \centering
    \vspace{-4mm}
    \includegraphics[width=\linewidth]{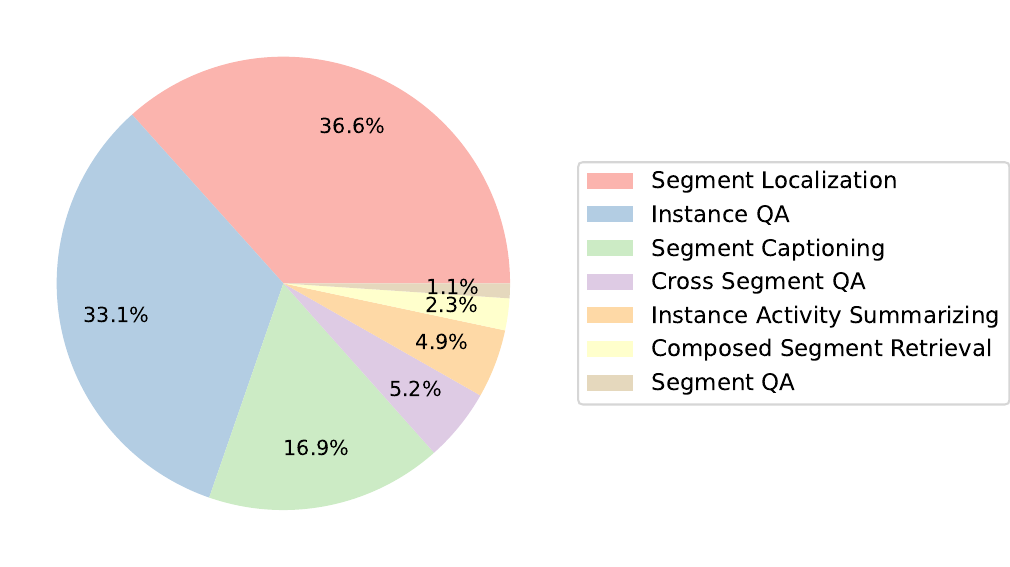}
    \vspace{-10mm}
    \caption{Distribution of different tasks in \texttt{Moment-10M}.}
    \vspace{-6mm}
    \label{fig:statistics}
\end{figure}

\vspace{-3mm}
\subsection{Instruction Generation}
\label{section:instruction_generation}
\vspace{-1mm}
We feed the information in the instance-event matrix into Vicuna \cite{vicuna}, an open-source text-based LLM, to generate instruction data. We design various types of instruction-following tasks to comprehensively train and evaluate Video-LLMs. We incorporate 5 tasks focusing on single segment understanding as well as 3 tasks that involve reasoning across multiple segments, as shown in Figure~\ref{fig:momentgen}. We utilize various prompts to guide Vicuna in generating instruction data for different tasks. Data from all 8 task types are used for instruction fine-tuning, while segment captioning data organized chronologically will be utilized for Grounded Event Sequence Modeling (Section~\ref{section:gesm}). Detailed task descriptions and prompts can be found in Appendix~\ref{section:task_formats} and \ref{section:prompts}. We select a substantial number of videos from YTTemporal-1B \cite{yttemporal1b} to build \texttt{Moment-10M}. Figure~\ref{fig:statistics} shows the distribution of each type of instruction data in \texttt{Moment-10M}. As shown in Table~\ref{tab:statistics}, \texttt{Moment-10M} comprises 10 million instruction data over 1.5 million segments and 451.5 thousand instance tracks. On average, each video contains 22.7 segments, which reflects the complexity of the event-sequences in the videos. We fine-tune \texttt{Momentor} on \texttt{Moment-10M}, enabling it to perform segment-level reasoning and localization.

\vspace{-1mm}
\section{Experiments}
\vspace{-1mm}
\subsection{Experiment Setup}
\vspace{-1mm}
To comprehensively evaluate \texttt{Momentor} in fine-grained understanding and precise localization, we assess it in a zero-shot setting across four tasks, i.e., \textit{action segmentation}, \textit{dense video captioning}, \textit{temporal grounding}, and \textit{highlight moment retrieval}, using datasets such as \mbox{Breakfast} \cite{breakfast}, \mbox{50 Salads} \cite{50salads}, \mbox{ActivityNet Captions} \cite{activitynet_captions}, \mbox{Charades-STA} \cite{charades_sta}, and \mbox{QVHighlights} \cite{qvhighlights}. We also perform evaluation on Video QA datasets such as \mbox{ActivityNet-QA} \cite{activitynet_qa}, \mbox{MSRVTT-QA}, and \mbox{MSVD-QA} \cite{msrvtt_msvd_qa} to evaluate \texttt{Momentor} in general question answering. Implementation details of \texttt{Momentor} can be found in Appendix~\ref{section:implementation}.

\begin{table}[t]
\centering
\resizebox{\linewidth}{!}{
\begin{tabular}{|c|cc|cc|cc|}
\hline
\multirow{3}{*}{Model} & \multicolumn{6}{c|}{Video QA} \\
\cline{2-7}
& \multicolumn{2}{c|}{MSVD-QA} & \multicolumn{2}{c|}{MSRVTT-QA} & \multicolumn{2}{c|}{ActivityNet-QA}\\
& Acc. & Score & Acc. & Score & Acc. & Score \\
\hline
Video-ChatGPT (7B) & 64.9 & 3.3 & 49.3 & 2.8 & 35.2 & 2.7 \\
VideoChat (7B) & 56.3 & 2.8 & 45.0 & 2.5 & 26.5 & 2.2 \\
Video-LLaMA (7B) & 51.6 & 2.5 & 29.6 & 1.8 & 12.4 & 1.1 \\
Valley (7B) & 65.4 & 3.4 & 51.1 & \textbf{3.0} & \textbf{45.1} & \textbf{3.2} \\
\textbf{\texttt{Momentor} (7B)} & \textbf{68.9} & \textbf{3.6} & \textbf{55.6} & \textbf{3.0} & 40.8 & \textbf{3.2} \\
\hline
\end{tabular}
}
\vspace{-3mm}
\caption{Existing Video-LLMs' performance on Video QA}
\vspace{-4mm}
\label{tab:video_qa}
\end{table}

\vspace{-3mm}
\subsection{Action Segmentation}
\vspace{-1mm}
Given a video, action segmentation requires the model to divide the video into multiple non-overlapping segments and assign an action category label to each segment. Since \texttt{Momentor}'s output is free-form text rather than action category labels, we use a sentence transformer \cite{sbert} to convert the output from \texttt{Momentor} into features, which are then compared with the features of action category labels to determine their corresponding action categories. We evaluate \texttt{Momentor} on \mbox{Breakfast} and \mbox{50 Salads}, of which the results can be referenced in Table~\ref{tab:exp_res1}. From the results we can infer: \textbf{(1)}~Overall, \texttt{Momentor} can effectively segment and recognize actions in input videos. In the setting of zero-shot action segmentation, \texttt{Momentor} achieves the highest accuracy among existing Video-LLMs. \textbf{(2)}~Despite only being trained on generating free-form texts rather than action labels, \texttt{Momentor}'s proficiency in visual information capturing still allows it to effectively generate texts that closely align with action label words, enabling it to perform accurate action classification.
\vspace{-1mm}
\subsection{Dense Video Captioning}
\vspace{-1mm}
Given a video, dense video captioning requires the model to output all events contained in the video along with their start and end timestamps. We test \texttt{Momentor} on \mbox{ActivityNet Captions}, and the results can be found in Table~\ref{tab:exp_res1}, from which we can conclude: \textbf{(1)}~Compared to existing Video-LLMs, \texttt{Momentor} provides more detailed event descriptions and more accurate event boundaries. \textbf{(2)}~Thanks to Grounded Event-Sequence Modeling, \texttt{Momentor} can capture the events in a video as completely as possible, while also providing precise start and end timestamps and accurate descriptions of each event. The model's leading performance just validates our viewpoint.
\vspace{-1mm}
\subsection{Temporal Grounding}
\vspace{-1mm}
Given a video and a natural language query, temporal grounding requires the model to identify the start and end timestamps of the segment corresponding to the query in the video. We evaluate \texttt{Momentor} on \mbox{ActivityNet Captions} and \mbox{Charades-STA}, with the results available in Table~\ref{tab:exp_res2}. Based on the experiment results, we can draw the following conclusions: \textbf{(1)}~\texttt{Momentor} achieves the highest mean IoU (Intersection over Union) among existing Video-LLMs. \textbf{(2)}~With the neighboring token propagation mechanism in the Temporal Perception Module, \texttt{Momentor} constructs a continuous and precise temporal token space, laying the foundation for accurate event localization. Ablation studies and visualization in \textbf{Section}~\ref{section:indepth_analysis} also validate this point. 
\subsection{Highlight Moment Retrieval}
\vspace{-1mm}
Given a video and a description of the highlight activities within the video, highlight moment retrieval requires the model to locate all the highlighted segments corresponding to the description. We evaluate \texttt{Momentor} on \mbox{QVHighlights}, and the results can be referenced in Table~\ref{tab:exp_res2}. From these results we can know: \textbf{(1)}~Among all existing Video-LLMs, \texttt{Momentor} achieves state-of-the-art performance on highlight moment retrieval. \textbf{(2)}~Thanks to the multi-event reasoning ability developed on the cross-segment tasks, \texttt{Momentor} can perceive the overall video semantics from a global perspective and effectively comprehend the relationships between different events, which is a key factor in highlight moment retrieval.

\begin{table}[t]
\centering
\resizebox{\linewidth}{!}{
\begin{tabular}{|c|c|c|c|c|}
\hline
\multirow{2}{*}{Setting} & \multicolumn{2}{c|}{ActivityNet} & Breakfast & QVHighlights \\
\cline{2-5}
& mIoU & CIDEr & MoF & mAP\\
\hline
\textbf{\texttt{Momentor} (7B)} & 29.3 & 14.6 & 24.4 & 7.6 \\ \hline
w/o CI & 27.6 & 13.1 & 22.5 & 7.1 \\
w/o NTP & 25.4 & 10.3 & 19.3 & 6.1 \\
w/o GESM & 27.8 & 9.8 & 19.5 & 6.8 \\
w/o Cross-Segment Tasks & 29.0 & 12.1 & 21.6 & 6.4 \\
\hline
\end{tabular}
}
\vspace{-3mm}
\caption{Performance of ablation models. CI: Continuous Interpolation, NTP: Neighboring Token Propagation, GESM: Grounded Event-Sequence Modeling}
\vspace{-2mm}
\label{tab:ablations}
\end{table}

\subsection{Video QA}
We test \texttt{Momentor} on ActivityNet-QA, MSRVTT-QA, and MSVD-QA. As shown in Table~\ref{tab:video_qa}, \texttt{Momentor} achieves state-of-the-art or comparative performance among Video-LLMs across all tested datasets, demonstrating its capability in coarse-grained video understanding.
\subsection{In-Depth Analysis}
\label{section:indepth_analysis}
\paragraph{Ablation Studies.} We conduct ablation experiments to assess the effectiveness of each component. The experiments are conducted under the following settings: (1)~w/o continuous interpolation: We still use temporal tokens to express temporal positions, but without integrating the continuous interpolation mechanism. (2)~w/o neighboring token propagation: We use the continuous temporal token space for temporal positioning, but without applying the neighboring token propagation mechanism when training. (3)~w/o grounded event-sequence modeling: After modality alignment, we proceed directly to instruction fine-tuning without grounded event-sequence modeling. (4)~w/o cross-segment tasks: We remove all instructions from cross-segment tasks and use only single-segment tasks for fine-tuning. We train \texttt{Momentor} with these settings and evaluate performances on ActivityNet Captions (temporal grounding and dense video captioning), Breakfast (action segmentation) and QVHighlights (highlight moment retrieval). The results of the ablation experiments can be referenced in Table~\ref{tab:ablations}.

\begin{figure}[t]
    \centering
    \includegraphics[width=\linewidth]{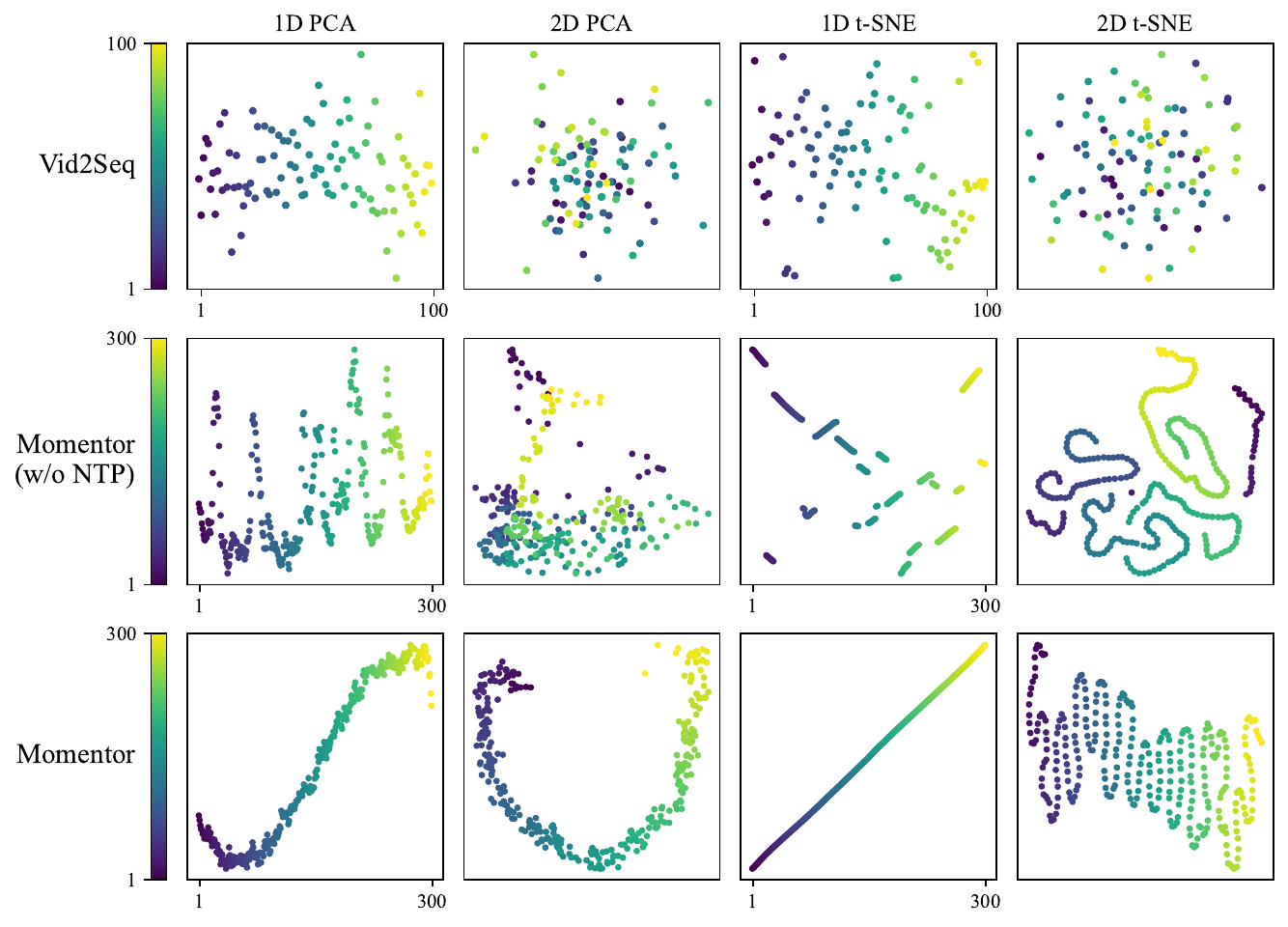}
    \vspace{-8mm}
    \caption{Visualization of temporal tokens in \texttt{Momentor} and time tokens in \textit{Vid2Seq}. NTP: neighboring token propagation.}
    \vspace{-4mm}
    \label{fig:visualization}
\end{figure}

Overall, removing any one of these components results in a decrease in the model’s overall performance. From Table~\ref{tab:ablations}, we can analyze the impact of removing different components on model performance separately. After removing the continuous interpolation mechanism, due to the quantization error, \texttt{Momentor} experiences a minor decline in localization-related metrics across all tasks, while the caption quality-related metrics are not significantly affected. Removing the neighboring token propagation mechanism leads to a performance drop in all metrics. Without neighboring token propagation, the temporal tokens are updated as multiple unrelated tokens rather than as an ordered sequence, which undermines the temporal representation and modeling. Visualizations of the temporal tokens (Section~\ref{fig:visualization}) also confirm this observation. Removing grounded event-sequence modeling leads to a significant performance decline in dense prediction tasks like dense video captioning and action segmentation, which indicates that grounded event-sequence modeling plays an important role in sequential semantics comprehension. The removal of cross-segment tasks has minimal impact on the performance of temporal grounding, as it does not involve cross-segment understanding. Performance on other tasks generally decreases, as both dense video captioning and action segmentation involve comprehension of multiple segments, and highlight moment retrieval also requires the model to distinguish between highlight segments and background segments.

\begin{figure}[t]
    \centering
    \begin{minipage}[b]{0.45\columnwidth}
        \includegraphics[width=\textwidth]{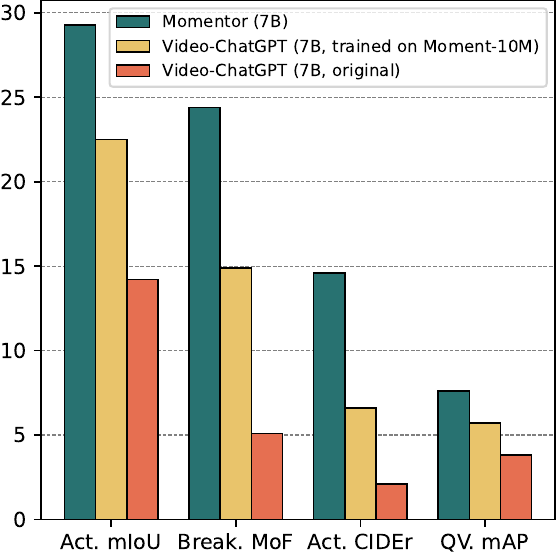}
        \vspace{-6mm}
        \caption{\small Dataset validation.\\Act.: ActivityNet.\\Break.: Breakfast.\\QV.: QVHighlights.}
        \label{fig:validation}
    \end{minipage}
    \hfill
    \begin{minipage}[b]{0.45\columnwidth}
        \includegraphics[width=\textwidth]{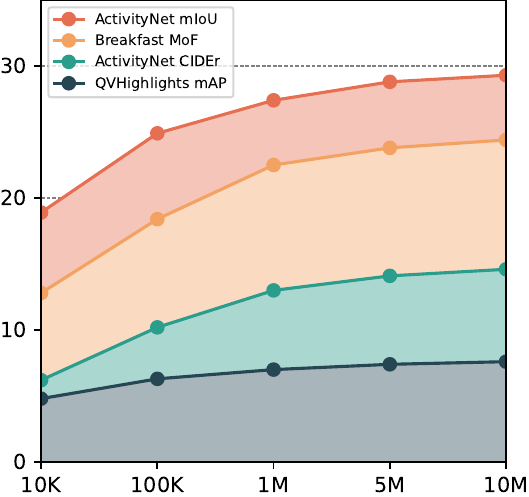}
        \vspace{-6mm}
        \caption{\small Impact of data scale. Generally, the performance improves as data scale increases.}
        \label{fig:scale}
    \end{minipage}
    \vspace{-4mm}
\end{figure}

\paragraph{Validation of \texttt{Moment-10M}.} We train Video-ChatGPT \cite{video_chatgpt} on our \texttt{Moment-10M} to validate its efficacy in improving fine-grained temporal reasoning. Despite being inefficient in temporal representation, we still use textual timestamps to represent temporal positions since Video-ChatGPT doesn't provide alternative temporal representation methods. As shown in Figure~\ref{fig:validation}, Video-ChatGPT trained on \texttt{Moment-10M} shows a great improvement on fine-grained temporal reasoning tasks.

\begin{figure}[t]
    \centering
    \includegraphics[width=\linewidth]{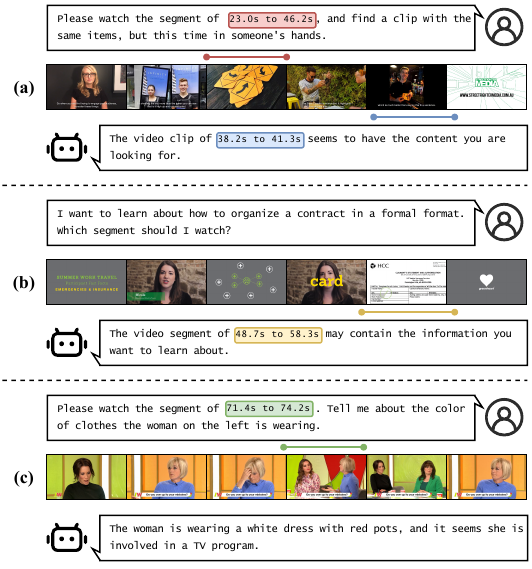}
    \vspace{-8mm}
    \caption{Analysis on special cases.}
    \vspace{-4mm}
    \label{fig:case_study}
\end{figure}

\paragraph{Impact of Data Scale.} We train \texttt{Momentor} with different amounts of instruction data, while the proportions of different tasks are kept the same. The results can be referenced in Figure~\ref{fig:scale}. Generally, the model's performance improves as the amount of training data increases, but slows down once the training data reaches a million-level scale.
\vspace{-3mm}
\paragraph{Case Studies.} We provide qualitative examples to demonstrate the fine-grained reasoning capability of \texttt{Momentor}. As shown in Figure~\ref{fig:case_study}(a), \texttt{Momentor} can integrate visual and textual input for comprehensive localization of target segment. Moreover, even when only a vague scene or requirement description is provided, \texttt{Momentor} can still understand the user's intent and pinpoint the segment containing relevant information, as exemplified in Figure~\ref{fig:case_study}(b). Additionally, although we don't incorporate spatial modeling, \texttt{Momentor} can still understand which instance the user is referring to and provide appropriate responses, as illustrated in Figure~\ref{fig:case_study}(c).
\vspace{-4mm}
\paragraph{Visualization of Temporal Tokens.} 
Since temporal tokens are used to represent uniformly distributed temporal positions, we expect them to exhibit continuity in their embeddings. We use PCA \cite{pca} and t-SNE \cite{tsne} to reduce the dimensionality of temporal tokens of \texttt{Momentor} and time tokens of \textit{Vid2Seq} \cite{vid2seq} into 1D and 2D for visualization. To validate the effectiveness of neighboring token propagation, we also visualize the temporal tokens trained without neighboring token propagation. For a fair comparison, we set the random state of t-SNE fixed to be 0. For the 1D reductions, we use the token indices as the x-axis and the reduced values as the y-axis; for the 2D reductions, we directly use the reduced values as coordinates. We employ a gradient color scheme, where the color of the data points will change progressively with the token index, as shown in Figure~\ref{fig:visualization}. It is evident that with neighboring token propagation, the embeddings of temporal tokens in \texttt{Momentor} are significantly more continuous. In contrast, embeddings of temporal tokens without neighboring token propagation and time tokens of \textit{Vid2Seq} exhibit much less continuity, as their correlation can only be learned indirectly and inefficiently.

\vspace{-3mm}
\section{Conclusion}
\vspace{-1mm}
We propose \texttt{Momentor}, a Video-LLM with segment-level comprehension and localization capabilities, and \texttt{Moment-10M}, a video instruction dataset comprising 10 million diversified instructions with segment-level annotation. We design a Temporal Perception Module to provide fine-grained temporal representation, and apply Grounded Event-Sequence Modeling to promote multi-event modeling in untrimmed videos. We train \texttt{Momentor} on \texttt{Moment-10M}, enabling it to perform comprehensive segment-level reasoning. Extensive experiments on various tasks demonstrate \texttt{Momentor}'s proficiency in fine-grained video understanding.

\vspace{-3mm}
\section*{Impact Statement}
\vspace{-1mm}
Our dataset, sourced from internet videos, is meticulously curated with stringent privacy safeguards. We acknowledge the potential presence of personal information and have instituted comprehensive measures to ensure its protection. Our model is conscientiously developed to be free from social harm and ethical breaches, embodying our commitment to responsible and beneficial technological advancement.

\vspace{-3mm}
\section*{Acknowledgements}
\vspace{-1mm}
This work was supported by the Key Research and Development Projects in Zhejiang Province (No. 2024C01106), the NSFC (No. 62372341), the National Key Research and Development Project of China (2018AAA0101900), Ant Group, and Research funding from FinVolution Group.

\bibliography{main}
\bibliographystyle{icml2024}

\newpage
\appendix
\onecolumn
\section{Overview}
In this appendix we present:
\begin{itemize}
    \item Implementation details of \texttt{Momentor} (Section~\ref{section:implementation}).
    \item Descriptions of the tasks in \texttt{Moment-10M} (Section~\ref{section:task_formats}).
    \item Prompts used for instruction generation (Section~\ref{section:prompts}).
\end{itemize}

\section{Implementation}
\label{section:implementation}
We utilize the CLIP \cite{clip} ViT-L/14 as the frame encoder and LLaMA \cite{llama} (7B) as the LLM. We initialize the linear projection layer with parameters from Video-ChatGPT's \cite{video_chatgpt} equivalent component. We incorporate $N=300$ temporal tokens for temporal positioning. For each video, we uniformly sample $M=300$ frames for fine-grained reasoning. We freeze the frame encoder and LLM during training, while only the linear projection layer and TPM are updated. We train \texttt{Momentor} on 8 A100 GPUs for around 60 hours. Our project is available at \href{https://github.com/DCDmllm/Momentor}{https://github.com/DCDmllm/Momentor}.

\section{Task Formats}
\label{section:task_formats}

\textbf{Single-Segment Tasks:}
\begin{itemize}
    \item \textit{Segment Captioning}: Given a segment, the Video-LLM is required to output a caption to conclude its content.
    \item \textit{Segment QA}: Given a segment, the Video-LLM is required to answer questions about that segment.
    \item \textit{Instance QA}: Given an instance at a certain moment, the Video-LLM is required to answer questions about that instance's behavior at that moment.
    \item \textit{Direct Segment Localization}: Given a query text, the Video-LLM is required to locate the described segment in the video and output its timestamp.
    \item \textit{Inferential Segment Localization}: Given a hypothetical scenario, the Video-LLM is required to find the scene in the video that likely correspond to that scenario and output its timestamp.
\end{itemize}

\textbf{Cross-Segment Tasks:}
\begin{itemize}
    \item \textit{Composed Segment Retrieval}: Given a source segment and the differences between the target and source segments, the Video-LLM is required to identify the target segment based on the source segment and these differences, and output its timestamp.
    \item \textit{Instance Activity Summarizing}: Given an instance, the Video-LLM is required to summarize the activities of this instance throughout the entire video.
    \item \textit{Cross-Segment QA}: Given multiple segments, the Video-LLM is required to combine information from all these segments to answer questions.
\end{itemize}
\newpage

\section{Prompts}
\label{section:prompts}
Below are the prompts used for generation of different kinds of instruction data. Due to page length constraints, we have omitted some in-context examples in certain tasks.

\begin{center}
\begin{tcolorbox}[colback=gray!20, colframe=black, text width=0.9\textwidth, title={Segment Captioning}]
Hello, I want you to act as a comprehensive video captioner. You will receive a list of frame-by-frame descriptions extracted from one video. Since some of these descriptions might be noisy, you should comprehend the major content of the video, assess the correctness of different pieces of information, and filter out erroneous, repetitive, noisy, or irrelevant details. After receiving and analyzing all the descriptions, please generate a comprehensive caption that effectively summarizes the events taking place in the video. Below are the information extracted from the video:\\
\\
\{descriptions\}\\
\\
There are some requirements that you should follow:\\
1. Your comprehensive video caption should be comprehensive, concise, informative, and LESS than 20 words.\\
2. You should output ONLY THE COMPREHENSIVE VIDEO CAPTION, and NO OTHER CONTENTS should be printed.\\
3. Your comprehensive video caption MUST NOT mention the concept of "frame" or "video".\\
Now please print out your comprehensive video caption.\\
\\
The comprehensive video caption:
\end{tcolorbox}
\end{center}

\begin{center}
\begin{tcolorbox}[colback=gray!20, colframe=black, text width=0.9\textwidth, title={Segment QA}]
Generate a concise dialogue with factual questions and their answers based on the following video segment caption:\\
\\
\{segment\_caption\}\\
\\
The answers should be directly inferred from the provided segment caption. Keep the questions and answers brief, with no more than 20 words each. Using "User" and "Assistant" as roles for questions and answers, respectively. Answer as if the "Assistant" can directly watch the video. Speak as a friendly and helpful assistant.
\end{tcolorbox}
\end{center}

\begin{center}
\begin{tcolorbox}[colback=gray!20, colframe=black, text width=0.9\textwidth, title={Instance QA}]
Generate a concise dialogue about the \{instance\_class\} with factual questions and their answers based on the following video segment caption:\\
\\
\{segment\_caption\}\\
\\
The answers should be directly inferred from the provided segment caption. Keep the questions and answers brief, with no more than 20 words each. Using "User" and "Assistant" as roles for questions and answers, respectively. Answer as if the "Assistant" can directly watch the video. Speak as a friendly and helpful assistant.
\end{tcolorbox}
\end{center}

\newpage

\begin{center}
\begin{tcolorbox}[colback=gray!20, colframe=black, text width=0.9\textwidth, title={Inferential Segment Localization}]
Hello, and I'd like you to act as a question generator. You will receive a sentence describing a clip in one video, and your task is to generate two questions with hypothetical scenario contexts to test if a deep learning model can retrieve the given clip based on the provided scenario information by asking about "what scene would you see" and "which clip should I watch" under that circumstance. Below are a few examples:\\
\\
\textless Example 1\textgreater \\
\\
\lbrack Clip Content\rbrack \\
\\
A elderly man is giving a speech in front of a blackboard.\\
\\
\lbrack Question\rbrack \\
\\
1. Suppose you are a college student and you are in class one morning. Which clip might demonstrate the scene in front of you at this time?\\
2. If I want to know how older generations teach classes, which clip should I watch?\\
\\
\textless Example 2\textgreater \\
\\
\lbrack Clip Content\rbrack \\
\\
A young woman is seen standing in a room and dancing around.\\
\\
\lbrack Question\rbrack \\
\\
1. You are a dance instructor. You are coaching your students in preparation for the next dance competition. What might you see at this moment? Please find the clip which might show this scene.\\
2. I want to learn to dance. Could you please tell me which clip of this video I should watch?\\
\\
\textless Example 3\textgreater \\
\\
\lbrack Clip Content\rbrack \\
\\
A dog in socks walks slowly out onto the floor as a lady films him.\\
\\
\lbrack Question\rbrack \\
\\
1. A female animal behavior researcher is studying the walking patterns of dogs when their feet tend to slip in your laboratory. Which clip in the video might you see at this point?\\
2. I'm feeling anxious and need to watch some funny animal videos to relax. Could you please help me find such a clip in the video?\\
\\
\\
Now given the following clip content sentence, please generate two questions with hypothetical scenario contexts to test if a deep learning model can retrieve the given clip based on the provided scenario information by asking about "what scene would you see" and "which clip should I watch" under that circumstance.\\
\\
\lbrack Clip Content\rbrack \\
\\
\{content\}\\
\end{tcolorbox}
\end{center}

\newpage

\begin{center}
\begin{tcolorbox}[colback=gray!20, colframe=black, text width=0.9\textwidth, title={Cross Segment QA}]
You are a visual assistant. Given several video clip descriptions, you are tasked to generate a concise factual question and its answer by combining information from all the video clip descriptions. Note that the \{instance\_class\} of all video clip descriptions in an input are the same, namely the video clip descriptions could describe the \{instance\_class\} at different time points. The answer should be directly inferred from the provided sentence. Keep the question and answer brief. Using "User" and "Assistant" as roles for questions and answers, respectively. Speak as a friendly and helpful assistant. Below are some examples:\\
\\
\textless Example 1\textgreater \\
\\
\lbrack Input\rbrack \\
\\
15.50s-30.75s : A group of children play in a park, running around and laughing.\\
45.20s-58.90s : A dog chases a frisbee, jumps to catch it, and returns it to its owner.\\
\\
\lbrack Output\rbrack \\
\\
User: What activities do the children and the dog engage in during the given video clips?\\
\\
Assistant: The children play in a park, running and laughing, while the dog chases a frisbee, jumps to catch it, and returns it to its owner.\\
\\
\textless Example 2\textgreater \\
\\
\lbrack Input\rbrack \\
\\
10.50s-30.88s : A chef in a white apron chops vegetables on a wooden cutting board.\\
50.20s-63.40s : A close-up of a sizzling steak on a hot grill.\\
80.16s-92.74s : A chef takes freshly baked bread out of the oven and places it on a cooling rack.\\
\\
\lbrack Output\rbrack \\
\\
User: What cooking activities can be observed in the video?\\
\\
Assistant: The video shows a chef chopping vegetables, frying steak on a hot grill, and taking freshly baked bread out of the oven.\\
\\
\\
Now given the following video clip descriptions, please generate a question-answer pair as it is in the examples. Note that the \{instance\_class\} of all video clip descriptions in an input are the same, namely the video clip descriptions may show the events or actions surrounding the \{instance\_class\} at different time points. The answer should be directly inferred from the provided sentence. Keep the question and answer brief. Using "User" and "Assistant" as roles for questions and answers, respectively. Speak as a friendly and helpful assistant.\\
\\
\lbrack Input\rbrack \\
\\
\{segment\_caption\}
\end{tcolorbox}
\end{center}

\newpage

\begin{center}
\begin{tcolorbox}[colback=gray!20, colframe=black, text width=0.9\textwidth, title={Instance Activity Summarizing}]
Hello, I want you to act as a comprehensive video captioner. You will receive a list of clip-by-clip descriptions extracted from one video. Since some of these descriptions might be noisy, you should comprehend the major content of the video, assess the correctness of different pieces of information, and filter out erroneous, repetitive, noisy, or irrelevant details. After receiving and analyzing all these descriptions, please generate a comprehensive caption that effectively summarizes the events taking place in the video about the \{instance\_class\}. Below are the clip descriptions extracted from the video:\\
\\
\{descriptions\}\\
\\
There are some requirements that you should follow:\\
1. Your comprehensive video caption should be comprehensive, concise, informative, and LESS than 20 words.\\
2. You should output ONLY THE COMPREHENSIVE VIDEO CAPTION, and NO OTHER CONTENTS should be printed.\\
3. Your comprehensive video caption MUST NOT mention the concept of "frame" or "video".\\
Now please print out your comprehensive video caption.\\
\\
The comprehensive video caption:
\end{tcolorbox}

\begin{tcolorbox}[colback=gray!20, colframe=black, text width=0.9\textwidth, title={Composed Retrieval}]
Hello, and I'd like you to act as a question generator. You will receive two descriptions about one source clip and one target clip. Your task is to generate one question with differences of the two clips to test if a deep learning model can retrieve the target clip based on the content of the source clip by asking about "could you please find a clip with following differences". Below are some examples:\\
\textless Example 1\textgreater \\
\\
\lbrack Source Clip Content\rbrack \\
\\
An old professor is giving a lecture in front of a blackboard.\\
\\
\lbrack Target Clip Content\rbrack \\
\\
An elderly man is giving a speech in front of a blackboard, holding a ruler.\\
\\
\lbrack Major Differences\rbrack \\
\\
The man in the target clip content is holding a ruler.\\
\\
\lbrack Instruction\rbrack \\
\\
Please watch the \{\{SOURCE\_CLIP\}\}. Could you please find a similar clip, but this time the speaker is holding something at hand?\\
\\
\textless Example 2\textgreater \\
\\
\lbrack Source Clip Content\rbrack \\
\\
A beautiful scene of primeval forest.\\
\\
\lbrack Target Clip Content\rbrack \\
\\
A beautiful view of coral reef taken in shallow sea.\\
\\
\lbrack Major Differences\rbrack \\
\\
The scene in the target clip is a seascape rather than a forest landscape.\\
\\
\lbrack Instruction\rbrack \\
\\
Please watch the \{\{SOURCE\_CLIP\}\}. Is there any similar clip with a different kind of scenery?\\
\\
Now given the following descriptions about one source clip and one target clip, please generate one question with differences of the two clips to test if a deep learning model can retrieve the target clip based on the content of the source clip by asking about "could you please find a clip with following differences". \\
\\
\lbrack Source Clip Content\rbrack \\
\\
\{source\_clip\_content\}\\
\\
\lbrack Target Clip Content\rbrack \\
\\
\{target\_clip\_content\}
\end{tcolorbox}
\end{center}



\end{document}